\documentclass[11pt,a4paper,svgnames]{article}
\usepackage{emnlp2021}
\usepackage{times}
\usepackage{latexsym}

\usepackage{adjustbox}
\usepackage{xcolor}
\usepackage{amsfonts}
\usepackage{amsmath}
\usepackage{amssymb}
\usepackage{amsthm}
\usepackage{stmaryrd}
\usepackage{xspace}
\usepackage{booktabs}
\usepackage{relsize}
\usepackage{comment}
\usepackage{multirow}

\usepackage{tikz}
\usetikzlibrary{arrows.meta}
\usetikzlibrary{shapes.arrows}
\usetikzlibrary{shapes.misc}
\usetikzlibrary{automata, arrows}
\usetikzlibrary{positioning}

\usepackage{inconsolata}

\usepackage{listings}
\usepackage{xcolor}

\definecolor{darkgreen}{rgb}{0.0,0.5,0.0}
\definecolor{darkblue}{rgb}{0.0,0.0,0.5}
\definecolor{aggblue}{rgb}{0,0,.65}
\definecolor{vargreen}{rgb}{0,.50,.18}

\newcommand{\var}[1]{{\texttt{{\color{vargreen} #1}}}\xspace}

\newcommand{\vI}{\var{I}}
\newcommand{\vJ}{\var{J}}
\newcommand{\vK}{\var{K}}
\newcommand{\vL}{\var{L}}

\newcommand{\vS}{\var{S}}

\newcommand{\vX}{\var{X}}
\newcommand{\vY}{\var{Y}}
\newcommand{\vZ}{\var{Z}}

\newcommand{\dd}[1]{{\small\tt #1}}

\lstdefinelanguage{dyna}{
  sensitive=true,
  morecomment=[l]{\%},
  morestring=[b]",
  classoffset=0,
  morekeywords={for,with\_key,arg,in,out},
  keywordstyle=\bf\tt\color{aggblue},
  classoffset=2,
  morekeywords={A,B,C,D,E,F,G,H,I,J,K,L,M,N,O,P,Q,R,S,T,U,V,W,X,Y,Z,Xs,Ys,Zs,DX,DY,DZ},
  keywordstyle=\color{vargreen},
  classoffset=0,
  keywordstyle=\color{blue},
  morekeywords={input,output},
}

\lstnewenvironment{dynaex}[2][]
                  {%
                    \lstset{
                      language={dyna},
                      morekeywords={for,out,with\_key,arg},
                      classoffset=2,
                      morekeywords={#2},
                      keywordstyle=\color{vargreen},
                      #1
                    }
                    \vspace{-2pt}
                  }
                  {%
                    \vspace{-2pt}
                  }

\usepackage{algorithm}
\usepackage[noend]{algpseudocode} 
\algrenewcommand\algorithmicindent{1.0em}%

\newcommand{\rightcomment}[1]{{\color{gray} \(\triangleright\) {\footnotesize\textit{#1}}}}
\algrenewcommand{\algorithmiccomment}[1]{\hfill \rightcomment{#1}}  
\algnewcommand{\LineComment}[1]{\State\rightcomment{#1}}
\algnewcommand{\LinesComment}[1]{\State\rightcomment{\parbox[t]{.95\linewidth-\leftmargin-\widthof{\(\triangleright\) }}{#1}}}

\algrenewcommand\alglinenumber[1]{{\tiny\color{black!50}#1.}\hspace{-2pt}}

\everypar{\looseness=-1}

\newcommand{\algorithmicfunc}[1]{\textbf{def} {#1}:}
\algdef{SE}[FUNC]{Func}{EndFunc}[1]{\algorithmicfunc{#1}}{}
\makeatletter
\ifthenelse{\equal{\ALG@noend}{t}}%
  {\algtext*{EndFunc}}
  {}%
\makeatother

\newcommand{\bigo}[1]{\mathcal{O}\!\left(#1\right)}
\newcommand{\mega}{\notation{\eta}}
\newcommand{\fold}[0]{\mysf{fold}}
\newcommand{\unfold}[0]{\mysf{unfold}}
\newcommand{\gensym}[0]{\mysf{gensym}}
\newcommand{\eliminate}[0]{\mysf{eliminate}}
\newcommand{\linearize}[0]{\mysf{linearize}}
\newcommand{\policy}[0]{\notation{\pi}}
\newcommand{\inputs}[0]{\mysf{inputs}}
\newcommand{\outputs}[0]{\mysf{outputs}}
\newcommand{\mcts}{\mysf{mcts}}

\usepackage{microtype}
\usepackage[inline]{enumitem}

\definecolor{darkgreen}{rgb}{0.0,0.5,0.0}
\definecolor{darkblue}{rgb}{0.0,0.0,0.5}
\definecolor{aggblue}{rgb}{0,0,.65}
\definecolor{vargreen}{rgb}{0,.50,.18}

\newcommand{\naive}[0]{na{\"i}ve\xspace}
\newcommand{\naively}[0]{na{\"i}vely\xspace}

\newcommand{\valpha}{\boldsymbol{\alpha}}
\newcommand{\vbeta}{\boldsymbol{\beta}}
\newcommand{\vtheta}{\boldsymbol{\theta}}

\newcommand{\defeq}[0]{\mathrel{\stackrel{\textnormal{\tiny def}}{=}}}

\newcommand{\dpluseq}{{\color{darkblue}\texttt{+=}}}
\newcommand{\dplus}{{\color{darkblue}\texttt{+}}}
\newcommand{\dtimes}{{\color{darkblue}\texttt{*}}}
\DeclareMathOperator*{\dprod}{{\color{darkblue}{\prod}}}

\newcommand{\tuple}[1]{\langle #1 \rangle}
\DeclareMathOperator*{\argmax}{\mathrm{argmax}}
\DeclareMathOperator*{\argmin}{\mathrm{argmin}}
\newcommand{\dopluseq}{{\color{darkblue}\oplus\texttt{=}}}
\newcommand{\doplus}{{\color{darkblue}\oplus}}
\newcommand{\dotimes}{{\color{darkblue}\otimes}}

\newcommand{\notation}[1]{{\color{aggblue}#1}}
\newcommand{\mysf}[1]{\notation{\textsf{\smaller #1}}}

\newcommand{\head}[1]{\mysf{head}{(#1)}}
\newcommand{\body}[1]{\mysf{body}{(#1)}}
\newcommand{\vars}[1]{\mysf{vars}{(#1)}}

\newcommand{\subst}[2]{\mysf{subst}(#1, #2)}
\newcommand{\fresh}[1]{\mysf{fresh}{(#1)}}
\newcommand{\Unify}[0]{\mysf{unify}}
\newcommand{\unify}[2]{\Unify(#1,#2)}

\newcommand{\cost}[0]{\mysf{cost}\xspace}
\newcommand{\transition}[0]{\mysf{transition}\xspace}

\newcommand{\degree}[0]{\mysf{degree}\xspace}
\newcommand{\InitialPolicy}[0]{\notation{\pi_0}}
\newcommand{\Update}[0]{\mysf{update}}
\newcommand{\vtof}[2]{\mysf{factors}(#1,#2)}
\newcommand{\elimvar}[1]{\mysf{elim}(#1)}

\newcommand{\Prog}[0]{\mathcal{P}}

\usepackage{calc}
\usepackage{thmtools}

\newtheorem{myexample}{Example}

\theoremstyle{definition}

\usepackage{cleveref}
\Crefname{ALC@unique}{Line}{Lines}
\crefname{section}{\S}{\S\S}
\Crefname{section}{\S}{\S\S}
\crefformat{section}{\S#2#1#3}
\crefname{table}{Table}{Tables}
\crefname{figure}{Fig.}{Fig.}
\crefname{algorithm}{Alg}{Alg}
\crefname{algorithm}{Alg}{Alg}
\crefname{line}{line}{lines}
\crefname{appendix}{App.}{App.}
\crefname{thm}{Theorem}{Theorems}
\crefname{myexample}{Example}{Examples}
\crefname{prop}{Proposition}{Propositions}
\crefname{defin}{Definition}{Definitions}
\crefname{lemma}{Lemma}{Lemmata}
\crefname{cor}{Corollary}{Corollaries}
\crefname{equation}{}{}

\newcommand{\defn}[1]{\textbf{#1}}

\title{Searching for More Efficient Dynamic Programs}

\usepackage{emoji}
\newcommand{\ucambridge}{\emoji[twitter]{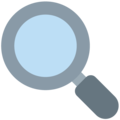}}
\newcommand{\ethz}{\emoji[twitter]{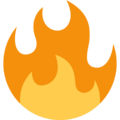}}
\newcommand{\jhu}{\emoji[twitter]{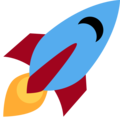}}

\author{
Tim Vieira\raise1.0ex\hbox{\normalfont\jhu}\raise1.0ex\hbox{\normalfont}~\;~Ryan Cotterell\raise1.0ex\hbox{\normalfont\ucambridge,\ethz}~\;~Jason Eisner\raise1.0ex\hbox{\normalfont\jhu}
\\
  \raise1.0ex\hbox{\normalfont\jhu}Johns Hopkins University \;
  \raise1.0ex\hbox{\normalfont\ucambridge}University of Cambridge \;
  \raise1.0ex\hbox{\normalfont\ethz}ETH Z\"{u}rich \\
  \href{mailto:tim.f.vieira@gmail.com}{\tt tim.f.vieira@gmail.com} \;\;
  \href{mailto:ryan.cotterell@inf.ethz.ch}{\tt ryan.cotterell@inf.ethz.ch}  \\
  \href{mailto:jason@cs.jhu.edu}{\tt jason@cs.jhu.edu}
}

\date{}

\begin{document}
\maketitle

\begin{abstract}
Computational models of human language often involve combinatorial problems. For instance, a probabilistic parser may marginalize over exponentially many trees to make predictions.  Algorithms for such problems often employ dynamic programming and are not always unique.
Finding one with optimal asymptotic runtime can be unintuitive, time-consuming, and error-prone. Our work aims to automate this laborious process.  Given an \emph{initial} correct declarative program, we search for a sequence of semantics-preserving transformations to improve its running time as much as possible. To this end, we describe a set of program transformations, a simple metric for assessing the efficiency of a transformed program, and a heuristic search procedure to improve this metric.  
We show that in practice, automated search---like the mental search performed by human programmers---can find substantial improvements to the initial program. Empirically, we show that many common speed-ups described in the NLP literature could have been discovered automatically by our system.
\end{abstract}

\section{Introduction}\label{sec:intro}
Algorithmic research in natural language processing (NLP) has focused---in large part---on developing dynamic programming solutions to combinatorial problems that arise in the field \cite{huang-2009-dynamic}.
Such algorithms have been introduced over the years for countless linguistic formalisms,
such as finite-state transduction \cite{mohri-1997-finite,eisner-2002-parameter,cotterell-etal-2014-stochastic},
context-free parsing \cite{stolcke-1995-efficient,goodman-1999-semiring},
dependency parsing \cite{eisner-1996-three,koo-collins-2010-efficient,ma-zhao-2012-fourth}
and mildly context-sensitive parsing \cite{vijay-shanker-weir-1989-recognition,vijay-shanker-weir-1990-polynomial,kuhlmann-etal-2018-complexity}.
In recent years, the same algorithms have often been used for deep structured prediction,
using a neural scoring function that decomposes over the structure \citep{durrett-klein-2015-neural,rastogi2016weighting,lee2016global,dozat2017biaffine,stern2017minimal,kim2017structured-attention,hong-huang-2018-linear,wu2018hard,wu-cotterell-2019-exact,qi2020stanza,rush-2020-torch-struct}.

When a dynamic programming algorithm for a new problem is first introduced in the literature, its runtime may not be optimal---faster versions are often published over time.
Indeed, the process of introducing a first algorithm and subsequently finding improvements is common throughout computer science.
In the case of dynamic programming, there are program transformations
that may be exploited to derive algorithms with a faster runtime \cite{eisner-blatz-2007}.  These transformations map a program to another program with the same meaning (given the same inputs, it will produce the same outputs), but with possibly different running time.  This paper shows how to search over program transformation sequences in order to automatically discover faster algorithms, automating the work of the NLP algorithmist.

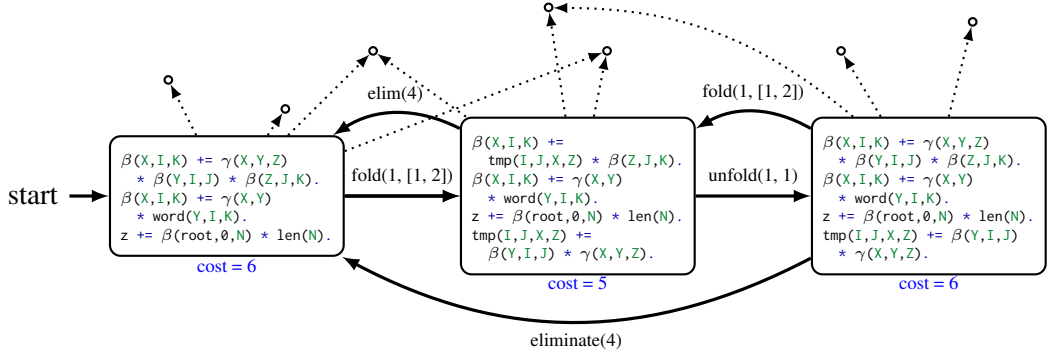
\begin{figure*}
\centering
\newcommand{\rarrow}[1]{\xrightarrow{\raisebox{-0.5ex}[0ex][0ex]{\tiny $#1$}}}

\definecolor{verydarkgray}{gray}{0.45}
\definecolor{darkgray}{gray}{0.6}

\begin{tikzpicture}[->,>=stealth',auto,node distance=3cm,
  thick,main node/.style={circle,draw,font=\sffamily\bfseries}]

\begin{scope}[>=latex,
          every edge/.style={draw, very thick}]

\tikzstyle{block} = [rectangle, draw, thick,
    rounded corners, inner sep=4pt, node distance={12em}, 
    font=\small\linespread{1}\selectfont]

\node [block, initial, minimum height=3.5em] (g1) at (0.8, 2.7) {
\begin{lstlisting}[basicstyle=\tiny\tt,language={dyna},numbers=none]
beta(X,I,K) += gamma(X,Y,Z)
  * beta(Y,I,J) * beta(Z,J,K).
beta(X,I,K) += gamma(X,Y) 
  * word(Y,I,K).
z += beta(root,0,N) * len(N).
\end{lstlisting}
};
    
\node[block, minimum height=4em] (g2) [right of=g1] {
\begin{lstlisting}[basicstyle=\tiny\tt,language={dyna},numbers=none]
beta(X,I,K) += 
  tmp(I,J,X,Z) * beta(Z,J,K).
beta(X,I,K) += gamma(X,Y) 
  * word(Y,I,K).
z += beta(root,0,N) * len(N).
tmp(I,J,X,Z) += 
  beta(Y,I,J) * gamma(X,Y,Z).
\end{lstlisting}
};

    \path [->] (g1) edge node[above] {\scriptsize fold(1, [1, 2])} (g2);

    \path [->] (g2) edge[bend right=30] node[above] {\scriptsize elim(4)} (g1);

\node [block] (g3) [right of=g2] {
\begin{lstlisting}[basicstyle=\tiny\tt,language={dyna},numbers=none]
beta(X,I,K) += gamma(X,Y,Z) 
  * beta(Y,I,J) * beta(Z,J,K).
beta(X,I,K) += gamma(X,Y) 
  * word(Y,I,K).
z += beta(root,0,N) * len(N).
tmp(I,J,X,Z) += beta(Y,I,J) 
  * gamma(X,Y,Z).
\end{lstlisting}
}; 

\begin{scope}[>=latex,
    every node/.style={node distance = 3em},
    every edge/.style={draw, dotted}]
\node (c1) [below of=g1,yshift=.6em] {\scriptsize\color{blue} cost = 6};
\node (c3) [below of=g2] {\scriptsize\color{blue} cost = 5};
\node (c4) [below of=g3] {\scriptsize\color{blue} cost = 6};
\end{scope}

    \path [->] (g1) edge (g2);
    \path [->] (g2) edge node[above] {\scriptsize unfold(1, 1)} (g3);    
    \path [->] (g3) edge[bend right=30] node[above] {\scriptsize fold(1, [1, 2])} (g2);
    
    \path [->] (g3) edge[bend left=28] node[below] {\scriptsize eliminate(4)} (g1);

\begin{scope}[>=latex,
    every node/.style={circle,thick,draw, inner sep=1pt, node distance = 4em},
    every edge/.style={draw, dotted}]
    
    \node (g4) [above of=g1, xshift = 2em, yshift=-1em] {};
    \node (g5) [above of=g1, xshift = -2em] {};
    \node (g9) [above of=g2, xshift = 1em, yshift=1em] {};
    \node (g40) [above of=g2, xshift = -1em, yshift=2.5em] {};
    \node (g23) [above of=g3, xshift = -3em, yshift=1em] {};
    \node (g24) [above of=g3, xshift = 1.5em, yshift=2em] {};
    \node (g27) [above of=g1, xshift = 5em, yshift=1em] {};

    \path [->] (g1) edge (g5);
    \path [->] (g1) edge (g27);
    \path [->] (g2) edge (g27);

    \path [->] (g3) edge[bend right=20] (g40);
    
    \path [->] (g1) edge (g4);
    \path [->] (g1) edge (g9);
    \path [->] (g2) edge (g9);
    \path [->] (g2) edge (g40);
    \path [->] (g3) edge[dotted] (g23);    
    \path [->] (g3) edge (g24);
    
\end{scope}

\end{scope}
\end{tikzpicture}
\caption{Depiction of the program optimization graph search problem (\cref{sec:search}).
The program used in this figure is our running example of speeding up CKY (\cref{ex:cky}).
Nodes are Dyna programs (\cref{sec:dyna}).
The node pointed to by ``start'' indicates the user's program.
Edges are program transformations (\cref{sec:transforms}).
Costs are derived by program analysis (\cref{sec:program-analysis}).
Only a tiny subset of the nodes and edges that exist in the search graph are shown.
The dotted unlabeled outgoing edges represent additional transformations that we did not elaborate in the diagram to reduce clutter.
}
\label{fig:search-space-diagram}
\end{figure*}

Consider the following instances\footnote{Many of these examples were brought to our attention in the works of \citet{eisner-blatz-2007} and \citet{gildea-2011-grammar}; further discussion can be found therein.} of published dynamic programs whose runtime bounds were later improved using specific applications of the program transformations mentioned above.
\begin{itemize}[leftmargin=1em, topsep=1pt, itemsep=1pt, parsep=1pt]
\item Projective dependency parsing: \newcite{collins-1996-new} gave an $\bigo{n^5}$ algorithm that was sped up to $\bigo{n^4}$ by \citet{eisner-satta-1999-efficient}.
\item Split-head-factored dependency parsing: implemented \naively runs in $\bigo{n^5}$; with some effort, an $\bigo{n^3}$ algorithm can be derived \cite{eisner-1996-three,johnson-2007-transforming,eisner-blatz-2007}.
\item Linear index-grammar parsing: $\bigo{n^7}$ in \citet{vijay-shanker-weir-1989-recognition},
sped up to $\bigo{n^6}$ by \citet{Vijay-Shanker93}.
\item Lexicalized tree adjoining grammar parsing: 
$\bigo{n^8}$ in \citet{vijay-shankar-joshi-1985-tag}, sped up to $\bigo{n^7}$ by \citet{eisner-satta-2000-faster}.
\item Inversion transduction grammar: $\bigo{n^7}$ in \citet{wu-1996-polynomial},
sped up to $\bigo{n^6}$ by \newcite{huang-etal-2005-machine}.
\item CKY parsing \cite{cocke-schwartz1970,younger67parsing,kasami65cky} is typically presented in a suboptimal $\bigo{K^3 n^3}$ form, but can be sped up to $\bigo{K^2 n^3 + K^3 n^2}$ \citep{lange2009cnf,eisner-blatz-2007}.
\item \citeauthor{10.5555/1623611.1623625}'s context-free parsing algorithm (\citeyear{10.5555/1623611.1623625}) runs in $\bigo{n^{\rho+1}}$ 
where $\rho$ is the length of the longest right-hand side of a context-free production in the grammar \cite{johnson-1989-computational}.
However, it can be made to run in $\bigo{n^3}$ by binarizing the production rules.
\end{itemize}

In this paper, we ask a simple question: Can we \emph{automatically} discover these faster algorithms?
Typically, a dynamic programming algorithm can be regarded as performing inference in an semiring-weighted deduction system \cite{goodman-1999-semiring}. \citet{eisner-goldlust-smith-2005} provided a programming language, Dyna, for expressing such deduction systems, along with a compiler that produced fast inference code.
All of the runtime improvements mentioned above are examples of source-to-source program transformations \cite{eisner-blatz-2007}.\footnote{The closest work in the NLP literature is
\citet{gildea-2011-grammar}, who proposed the junction-tree minimization to speed up dynamic programs, which corresponds to only considering the fold transformation. Outside of NLP, \citet{mastria202asp} learn to fold in the context of answer set programs.
Our work considers a broader range of program transformations.}\looseness=-1

Our work, depicted in \cref{fig:search-space-diagram}, poses program optimization as a search over transformed versions of the initial program, an idea that was suggested as future work by \citet{eisner-blatz-2007}.
Our contribution is to show that two classic search algorithms---beam search \cite{reddy-1977,meister-etal-2020-best} and Monte Carlo tree search \cite{kocsis2006bandit}---are effective for this purpose, rapidly rediscovering many of the known optimizations listed above.
To set up this solution, the following sections describe the elements of the search problem: 
our space of possible programs (\cref{sec:dyna}), 
a simple cost function that serves as a proxy for program runtime (\cref{sec:program-analysis}), 
and a set of directed edges that connect semantically equivalent programs (\cref{sec:transforms}). 
Our search starts at the initial program and seeks a low-cost equivalent program that can be reached by traversing directed edges.  In \cref{sec:search}, we review the beam search and MCTS algorithms that we will use for this purpose in the experiments of \cref{sec:experiments}.\looseness=-1

\section{Our Space of Dynamic Programs}
\label{sec:dyna}

We will consider programs that are expressed in the original version of the Dyna language \cite{eisner-goldlust-smith-2005},
which is essentially a way of writing down the recurrence relations of a dynamic programming algorithm.
In this section, we only briefly describe the language and refer the reader to \citet{eisner-goldlust-smith-2005} for a more complete introduction.
To start, consider the following examples.
\begin{myexample}
\label{ex:z:dyna}
The total weight of length-4 paths in a graph with edge weights \dd{w}:
\begin{dynaex}{}
@\hspace{-4pt}@z += w(Y1,Y2) * w(Y2,Y3) * w(Y3,Y4) * w(Y4,Y5).
\end{dynaex}
\end{myexample}
\noindent This program defines the value of a derived \defn{item} \dd{z} in terms of input items \dd{w(...)}.  The value of \dd{z} is
$\sum_{\dd{\vY_1}} \!\cdots\! \sum_{\dd{\vY_5}} \scriptstyle \dd{w(\vY_1,\vY_2) \cdot w(\vY_2,\vY_3) \cdot w(\vY_3,\vY_4) \cdot w(\vY_4,\vY_5) }$.\footnote{We chose the name \dd{z} as it is short and traditional for denoting the normalization constant of a probabilistic model, e.g., $\dd{p(\vY_1,\vY_2,\vY_3,\vY_4,\vY_5) \propto w(\vY_1,\vY_2) \!\cdot\! w(\vY_2,\vY_3) \!\cdot\! w(\vY_3,\vY_4) \!\cdot\! w(\vY_4,\vY_5)}$ would be have \dd{z} as its normalization constant.}
Notice that there is no need to clutter the expression with explicit summations
or control-flow constructs such as for-loops: all variables (denoted by capitalized letters) that appear only on the right-hand side of $\dpluseq$ are summed over.

Dyna programs elegantly enable recursive computation by allowing the value of an item to be defined in terms of other items of the same kind.
\begin{myexample}
\label{ex:viterbi}
The Viterbi algorithm \citep{viterbi67error} finds the most probable path in a graph from a node labeled \dd{"a"} to a node labeled \dd{"z"} with edge probabilities (or weights) \dd{w}:
\begin{dynaex}{}
alpha("a") max= 1.
alpha(J) max= alpha(I) * w(I,J).
z += alpha("z").
\end{dynaex}
\end{myexample}
\noindent The second rule is recursive.  For each possible value of $\vJ$ on the left-hand side, it defines \dd{$\alpha$(\vJ)} by maximizing over assignments to the \emph{other} variables on the right-hand side (namely $\vI$).  Maximization is specified by \dd{max=}, whereas summation in \cref{ex:z:dyna} was specified by \dd{+=}.

\begin{myexample}\label{ex:cky}
Weighted context-free parsing with CKY \citep{cocke-schwartz1970,younger67parsing,kasami65cky}, or more precisely the inside algorithm \citep{baker79trainable,jelinek1985markov}:\footnote{If the reader is not familiar with context-free parsing, we recommend \citet[chapters 12--13]{jurafsky-martin-book}.}
\begin{dynaex}{}
@\!\!@beta(X,I,K) += beta(Y,I,J) * beta(Z,J,K) * gamma(X,Y,Z).
@\!\!@beta(X,I,K) += gamma(X,Y) * word(Y,I,K).
@\!\!@z += @$\beta$@(root,0,N) * len(N).
\end{dynaex}
\end{myexample}
\noindent The values of the $\gamma$ items should be defined to be the weights of the corresponding context-free grammar rules: for example, the item \dd{$\gamma$(s,np,vp)}\,=\,$0.7$ encodes
the production $\dd{s} \xrightarrow{\scriptsize 0.7} \dd{np}\, \dd{vp}$).  Also, the item $\dd{word(\vX,\vI,\vK)}$ should be 1 if the input word $\vX$ appears at position $\vI$ of the input sentence and $\vK=\vI+1$, and should be 0 otherwise.
Then for any nonterminal symbol $\vX$ and any substring spanning positions [\vI, \vK) of the input sentence, the item $\beta$\dd{(\vX,\vI,\vK)}  
represents the total weight of all grammatical derivations of that substring from $\vX$. 

\newcommand{\db}[0]{\dd{b}}
\newcommand{\dhead}[0]{\dd{h}}

More generally, a Dyna \defn{program} $\Prog$ is a collection of \defn{rules}, each rule having the form $\dhead \;\dopluseq\; \db_1 \,\dotimes \cdots \dotimes\, \db_K$.  Here $\tuple{\doplus, \dotimes}$ can be any pair of operations that form a \defn{semiring} \citep{goodman-1999-semiring,huang-2009-dynamic}, such as $\langle \dplus, \dtimes \rangle$ in \cref{ex:z:dyna} and \cref{ex:cky}, or $\langle{\color{aggblue}\texttt{max}}, \dtimes\rangle$  in \cref{ex:viterbi}.\footnote{Many other semirings are useful in NLP \cite{goodman-1999-semiring,huang-2009-dynamic,eisner-goldlust-smith-2005}.}
We call $\dhead$ the head, and $\db_1, \cdots, \db_K$ the body of the rule.  
Each $\db_k$ in the body is called a \defn{subgoal}.  
Let $\head{r}$ and $\body{r}$ denote the head and body terms in a rule $r$.  
We assume that all rules in the program use the same semiring.\footnote{We leave the extensions in Dyna 2 \cite{dyna2}, which relaxes this restriction, to future work.}
The structured names of items are \defn{terms}, which are nested typed tuples as in Prolog.  For example, \dd{f(g(z,h(3)))} is a 1-tuple of type \dd{f}, whose single element is a 2-tuple of type \dd{g}, and so on.  The rules use captialized variables such as $\vX$ to pattern-match against subterms, where a variable that is repeated in a rule must have the same value each time. 
Let $\vars{\cdot}$ denote the set of variables contained in a term, e.g., $\vars{\dd{f(g(\vX),4,\vX)}} \mapsto \{\vX\}$.
The Dyna language allows logical \defn{side conditions} on a rule, e.g., \dd{goal += f(\vX) for \vX < 10.} This is syntactic sugar for \dd{goal += f(\vX) * lessthan(\vX,10)}, where the value of each \dd{lessthan($a$,$b$)} term is the one or zero element of the semiring, according to whether $a < b$ or not.

\section{Program Analysis}
\label{sec:program-analysis}

Our goal is to search for a \emph{fast} Dyna program.  We will assume that the programs are executed using the forward chaining algorithm described by \citet{eisner-goldlust-smith-2005}.\footnote{This algorithm assumes that the program is range-restricted, i.e., $\vars{\head{r}} \subseteq \vars{\body{r}}$.\label{def:range-restricted}  An example of a non-range restricted rule is \dd{id(\vI,\vI) += 1}.}
In principle, we could evaluate a candidate program's runtime by actually executing it, but this would be very expensive and would also require us to specify particular inputs to the program.  Instead, as our search objective, we will use a simple asymptotic upper bound on the program's runtime, based on a folk theorem from the Datalog community that has a long history of use. Many NLP papers have analyzed the runtime of their algorithms using either this folk theorem or a more refined version given by \citet{mcallester-2002-meta}: \citet{gildea-2011-grammar,nederhof-satta-2011-prefix-probability,gilroy-etal-2017-parsing,melamed-2003-multitext,kuhlmann-2013-mildly,nederhof-sanchez-saez-2011-parsing,buchse-etal-2011-tree,lopez-2009-translation,eisner-blatz-2007}.

The folk theorem says that, under certain conditions (discussed later), the running time of forward chaining execution of a given program is at worst linear in the number of ways to instantiate its rules, i.e., bind the variables to constants.  A relatively simple bound on rule instantiations is available if we can establish that each variable in the program can be bound in at most $\mega$ different ways.  In that case, given some other conditions discussed at the end of this section, the number of ways to instantiate a rule with $k$ variables is bounded by $\bigo{\mega^k}$.  Program $\Prog$'s total runtime is $\bigo{\mega^{\degree(\Prog)}}$ where $\degree(\Prog)$ is the maximum number of variables in any rule of $\Prog$.  We therefore take $\degree$ to be the cost function to minimize during search.

Consider \cref{ex:z:dyna}. Evaluating this program under the forward-chaining algorithm will instantiate the rule by binding the 5 variables $\vX_1, \vX_2, \vX_3, \vX_4, \vX_5$ to constants.  Then, the number of rule instantiations is $\bigo{\mega^5}$.

Similarly, \cref{ex:cky} runs in $\bigo{\mega^6}$, as the first rule must sum over 6 variables, $\vX, \vY, \vZ, \vI, \vJ, \vK$.
Note that this is a coarse-grained analysis: the runtime is usually given more specifically as $\bigo{n^3 K^3}$ where $n$ is the number of sentence positions, and $K$ is the number of grammar symbols.  (This finer-grained bound can be achieved by the theorem of \citet{mcallester-2002-meta}: the intuition is that the variables $\vI, \vJ, \vK$ can each be bound in $\bigo{n}$ ways while $\vX, \vY, \vZ$ can each be bound in $\bigo{K}$ ways.)  However, the simpler analysis $\bigo{\mega^6}$ gives us a single exponent to reduce, namely 6.

To see the cost function in action, consider that \cref{ex:z:dyna} has a running time of $\bigo{\mega^5}$, whereas the following equivalent program runs in $\bigo{\mega^2}$.
\begin{myexample}
\label{ex:z:good-fold}
Efficient factorization of \cref{ex:z:dyna}
\begin{dynaex}{}
z += rest@$_1$@(Y@$_1$@).
rest@$_1$@(Y@$_1$@) += w@$_1$@(Y@$_1$@,Y@$_2$@) * rest@$_2$@(Y@$_2$@).
rest@$_2$@(Y@$_2$@) += w@$_2$@(Y@$_2$@,Y@$_3$@) * rest@$_3$@(Y@$_3$@).
rest@$_3$@(Y@$_3$@) += w@$_3$@(Y@$_3$@,Y@$_4$@) * rest@$_4$@(Y@$_4$@).
rest@$_4$@(Y@$_4$@) += w@$_4$@(Y@$_4$@,Y@$_5$@) * rest@$_5$@(Y@$_5$@).
\end{dynaex}
\end{myexample}

Similarly, in \cref{ex:cky}, we can sum over the variable \vY separately from \vK as follows:
\begin{myexample}{}\label{ex:cky-fold}
Faster CKY (\cref{ex:cky})
\begin{dynaex}{}
@$\beta$@(X,I,K) += tmp(I,J,X,Z) * @$\beta$@(Z,J,K).
tmp(I,J,X,Z) += @$\beta$@(Y,I,J) * @$\gamma$@(X,Y,Z).
\end{dynaex}
\end{myexample}
\noindent which is more efficient as its running time is $\bigo{\mega^5}$.  It is also more efficient under the finer-grained analysis, $\bigo{K^2 n^3 + K^3 n^2}$.

The \degree analysis of a Dyna program only leads to a \emph{valid} $\mathcal{O}$-expression under some conditions, which we will now discuss.
(1) The \degree bound requires the \emph{grounded} program to be \defn{acyclic} \citep{eisner-goldlust-smith-2005}.  Cycles slow down forward chaining because it must iterate to a numerical fixed point. Generally, the number of iterations required to reach a fixed-point is data dependent.\footnote{
In the Boolean case (i.e., a Datalog program), the cycles do not affect the running time because finding ``new'' values for an item that is already true does not trigger further propagation to items that depend on it.  Unfortunately, this is not true of general semirings.  For example, the program \dd{a += r * a. a += 1.} encodes a geometric series; it may take many iterations to converge if $|\dd{r}|$ is close to $1$, and will diverge if $|\dd{r}| \ge 1$. The efficiency of this cyclic program thus depends on the value of the input parameter \dd{r}.}
(2) The \degree bound assumes that all of the relations in the program are \defn{bounded} in size.  The \degree bound requires that terms are not nested;  this prevents the user from encoding infinite sets, such as the Peano integers. Additionally, it assumes that the program's rules are all range-restricted (\cref{def:range-restricted}). (3) The \degree bound also assumes that the semiring operations are constant time.

We will see in  \cref{sec:experiments} that simply optimizing \degree is sufficient to recover a number of asymptotic speedups noted in the NLP literature (see \cref{sec:intro}) as well as asymptotic speedups on synthetic programs.

That said, the \degree analysis might be loose for many reasons.  The upper bounds derived using our methodology assume that relations are dense.  Often relations are statically known to be sparse.
Many low-level details affect actual execution time, but do not matter for asymptotic complexity.
For example, memory layouts (e.g., row-order or column-order layout of a dense array in memory), sparse vs. dense representations of relations (e.g., hash tables vs. arrays), and indexes on relations (including sorted order) can have a dramatic effect on the running time in practice. However, they will not manifest in the \degree analysis (e.g., \citet{PHIPAC-1997,dunlop-2011-efficient,denero-etal-2009-efficient,lopez2007hierarchical}).
Such choices are out of the control of our specific search space,
but they may interact with the program in ways that are not represented in the \degree.

An obvious alternative cost function would be the empirical execution time of executing the transformed program on a workload of representative inputs (e.g., running a transformed parser on actual sentences from the Penn Treebank \citep{ptb}).  But as we noted earlier, such a cost function might be impractically expensive.
For example, evaluating the \degree of a degree-1000 program is linear in the size of the program, whereas evaluating the wallclock time is $\bigo{\mega^{1000}}$. Optimizing the program degree is a crucial design choice as it enables a more exhaustive search in practice.  Additionally, it sidesteps the need to optimize for a specific workload.
However, in future work, we would like to investigate hybrid search algorithms (e.g., \citet{song2019mfbo}) that do attempt to minimize empirical execution time, but replace some of the expensive evaluations of that with cheaper approximations: empirical execution time but with a timeout, fine-grained bounds obtained by abstract interpretation, and---most cheaply---estimates derived from worst-case asymptotic analysis as above.

\section{Program Transformations}
\label{sec:transforms}
This section details the set of transforms that we consider in this paper.  None of the transforms are novel to this work.  They have been detailed and proved correct in \citet{eisner-blatz-2007}.  We include them in our discussion for completeness of presentation and to illuminate the challenges of our search problem. We provide pseudocode for the transforms in \cref{app:transform-pseudocode}, but defer to \citet{eisner-blatz-2007} for a more thorough discussion.

\paragraph{Input and output declarations.}
We assume that the initial program declares some items as input and/or output items.  The rest are considered intermediate items.  A program transform must preserve the mapping from a valuation of the input items to a valuation of the output items.  (A valuation is an assignment of a value to each item.)  However, a program transform is free to introduce, destroy, or alter intermediate items.

For example, for CKY, the input and output items are declared as  follows:
\begin{dynaex}{}
input word(X,I,K); @$\gamma$@(X,Y,Z); @$\gamma$@(X,Y).
output goal.
\end{dynaex}

\subsection{Fold}
\label{sec:fold}

\Crefrange{ex:z:dyna}{ex:cky} were examples of the folding transform.
Our candidate fold actions are based on \defn{variable elimination}, as these are the only valid folding actions that reduce the rule's \degree.  For a given rule $r$, the variable $v \in \vars{r}$ \emph{can be eliminated} if it does not appear in \emph{all} of the factors in the rule's body and it does not appear in the head of the rule.  Formally, the set of such variables is $\elimvar{r} \defeq \{v \mid \vtof{r}{v} \ne \body{r}, v \notin \head{r} \}$ where $\vtof{r}{v} \defeq \{ b \mid b \in \body{r}, v \in \vars{b} \}$.

If any rule $r$ of the program $\Prog$ contains variables that can be eliminated ($|\elimvar{r}| > 0$), then eliminating any variable $v \in \elimvar{r}$ by folding $\vtof{r}{v}$ out of $r$ reduces that $r$'s \degree, which may reduce (and never increases) the \degree of the program.  Therefore, no final program benefits from having rules with variables that can be eliminated.
However, when more than one variable can be eliminated ($|\elimvar{r}|>1$), the order in which the variables are eliminated will affect the eventual \degree.  Finding an optimal sequence of variable-elimination steps is NP-hard, by reduction from variable elimination ordering in probabilistic graphical models \citep{gildea-2011-grammar}.

We briefly note that folding can increase the space complexity of the program, since it introduces intermediate items that will be stored.  We do not consider optimizing the space--time tradeoff, but it could be done with methods similar to ours.

\subsection{Unfold and Rule Elimination}
\label{sec:unfold}
\label{sec:rule-elimination}

Suppose the user provided an inefficient program, such as \cref{ex:z:bad-fold}, which could have been obtained by folding \cref{ex:z:dyna} with a suboptimal variable-elimination ordering.

\begin{myexample}
\label{ex:z:bad-fold}
Bad ordering for \cref{ex:z:dyna}
\begin{dynaex}{}
goal += tmp@$_1$@(X@$_1$@,X@$_4$@,X@$_5$@).
tmp@$_4$@(X@$_1$@,X@$_2$@) += w@$_1$@(X@$_1$@,X@$_2$@).
tmp@$_3$@(X@$_2$@,X@$_4$@) += w@$_2$@(X@$_2$@,X@$_3$@) * w@$_3$@(X@$_3$@,X@$_4$@).
tmp@$_1$@(X@$_1$@,X@$_4$@,X@$_5$@) += tmp@$_2$@(X@$_1$@,X@$_4$@) * w@$_4$@(X@$_4$@,X@$_5$@).
tmp@$_2$@(X@$_1$@,X@$_4$@) += tmp@$_4$@(X@$_1$@,X@$_2$@) * tmp@$_3$@(X@$_2$@,X@$_4$@).
\end{dynaex}
\end{myexample}

While the above program is correct, its \degree is 3, which is worse than the optimal variant, which has \degree 2.  It has no variables that can be eliminated, so there are no fold actions that can improve its degree.  To improve it, we first have to \emph{undo} the poor choices.
There are two transformations for ``undoing'' folds: unfold and rule elimination, which we will describe in this section.

The \defn{unfold} transform is essentially the inverse of the fold transformation, and corresponds to inlining code.  It takes as input a specific subgoal $\dd{b_k} \!\in\! \body{r}$ of some rule $r$.  The goal is to replace $\dd{b_k}$ by its definition.  We will remove $r$ from the program, and replace it by adding a specialized version of $r$ for each rule $r'$ whose head unifies with $\dd{b_k}$.  These rules $r'$ define $\dd{b_k}$---except in the special case where $\dd{b_k}$ matches any input items, in which case we cannot unfold $\dd{b_k}$ because its complete definition is not available.

As a simple example, consider unfolding the first subgoal of the first rule of \cref{ex:cky-fold},
\begin{dynaex}{}
@$\beta$@(X,I,K) += tmp(I,J,X,Z) * @$\beta$@(Z,J,K).
tmp(I,J,X,Z) += @$\beta$@(Y,I,J) * @$\gamma$@(X,Y,Z).
\end{dynaex}
This becomes
\begin{dynaex}{}
@$\beta$@(X,I,K) += @$\beta$@(Y,I,J) * @$\gamma$@(X,Y,Z) * @$\beta$@(Z,J,K).
tmp(I,J,X,Z) += @$\beta$@(Y,I,J) * @$\gamma$@(X,Y,Z).
\end{dynaex}
Notice that the second rule is now defunct.\footnote{We make use of a simple dead rule detection strategy to identify rules that cannot fire based on the declared $\inputs$, or are unused by any of the declared $\outputs$. Determining which rules are dead is possible with a straightforward graph reachability analysis on a coarsened program, such as the program resulting from dropping the arguments to all relations, known as the predicate graph.}  This transformation is correct by the distributive rule.  Notice that the \degree increased from 5 to 6.

An unfold will usually increase or preserve the program's \degree---but there are exceptions due to repeated variables or constants in rules:
\begin{dynaex}{}
a(I,K) += b(I,J) * c(J,K).
trace += a(L,L).
\end{dynaex}
Unfolding the subgoal of the second rule decreases the program's degree from 3 to 2:
\begin{dynaex}{}
trace += b(L,J) * c(J,L).
\end{dynaex}
Thus, this is an example of an immediately useful unfold.  The program in this case computes the trace of a matrix product, and the role of unfold is to \emph{specialize} the sub-program that computes the entire matrix product \dd{a(\vI,\vK)} to the site where the product is used, which only seeks its diagonal \dd{a(\vL,\vL)}.  Such optimizations are easy for programmers to miss.

\defn{Rule elimination} is an alternative transformation, also described by \citet{eisner-blatz-2007}, that happens to achieve the same result in the above examples. As the name suggests, rule elimination targets rules instead of subgoals. The transform takes as input a rule $r'$, removes it from the program, and adds specialized versions of all of the rules $r$ whose subgoals match the head of $r'$.  Rule $r'$ cannot be eliminated if its head matches any output items, since that would change the value of those output items.  Notice that attempting to eliminate recursive rules is futile, as a rule cannot be eliminated until it reaches its base case.

Rule elimination and unfold are especially useful for eliminating non-range-restricted rules (\cref{def:range-restricted}).  For example, eliminating the first rule from

\begin{dynaex}{}
f(I) += 1.
f(I) += g(I) * m(I,J).
goal += f(I) * h(I).
\end{dynaex}
yields the range-restricted program
\begin{dynaex}{}
f(I) += g(I) * m(I,J).
goal += f(I) * h(I).
goal += 1 * h(I).
\end{dynaex}

Another useful case of rule elimination is for propagating constants throughout the program.  For example, if the grammar in \cref{ex:cky} is known in advance, i.e., $\dd{\gamma(\vX,\vY,\vZ)} \notin \inputs$, then we can propagate them ahead of time.  Yielding a highly specialized program with no \vX, \vY, \vZ variables and having an overall reduced \degree of 3.

In order to recover the original version of \cref{ex:z:dyna} given \cref{ex:z:bad-fold}, we can eliminate all rules except for the one defining the output item \dd{z}, and then fold to eliminate variables in a different order.  This poses search challenges because all of the unfold or rule elimination actions needed to reach \cref{ex:z:dyna} are ``uphill'': they increase the \degree, until the folds are applied.  This means that finding useful unfold and rule elimination moves can be challenging (i.e., take a fair amount of exploration).

\section{Program Improvement}
\label{sec:search}

Our goal is to find a sequence of transformations to the user's program $\mathcal{P}_0$ that gives the lowest cost, $\cost(\Prog) \defeq \degree(\Prog)$.  In this section, we provide two effective search algorithms for approaching this goal: beam search and Monte Carlo tree search.\looseness=-1

\subsection{The Graph Search Problem}

We consider an abstract \defn{graph search problem}, $\langle \mathcal{S}, \mathcal{A}, s_0, \mathcal{T}, \transition, \cost\rangle$,
where
$\mathcal{S}$ is a state space,
$\mathcal{A}$ is an action space,
$\transition: \mathcal{S} \times \mathcal{A} \to \mathcal{S}$ is a transition function,
$s_0 \in \mathcal{S}$ is an initial state,
$\mathcal{T} \subseteq \mathcal{S}$ is a set of terminal states,
and $\cost: \mathcal{T} \to \mathbb{R}_{\ge 0}$ is an cost function on the terminal states.  The $\cost$ and $\transition$ functions will be treated by MCTS and beam search as black boxes.\footnote{MCTS can be used in the more general case where $\cost$ and $\transition$ are stochastic or adversarial functions, but this is not the case in our setting.}  The goal of the search problem is to find the terminal state in the graph that has the lowest cost, $s^* = \min_{s \in \mathcal{T}} \cost(s)$.

Our problem (depicted in \cref{fig:search-space-diagram}) can be easily mapped into this notation.
Our states $\mathcal{S}$ are programs (\cref{sec:dyna}).
The initial state $s_0$ is the initial program $\Prog_0$.
The transitions are applications of any valid program transformation, which we discussed in \cref{sec:transforms}.  
In our setting, every state is a terminal.
For the cost function, we use program's $\degree$ (\cref{sec:program-analysis}).
To ensure termination, we only explore up to a distance of 100 from the initial state.
We will discuss in \cref{sec:search-refined} how to structure the state and action spaces to make search more effective.

\subsection{Beam Search}
Beam search is a common heuristic search algorithm, 
which is easy to implement (\cref{alg:beam-search}) and often works well in practice. 
A terse description of the algorithm is that it is a variant of breadth-first search \cite{russell-norvig} which prunes the search frontier (FIFO queue) to only keep the $B$ lowest $\cost$ states so far.\footnote{
In our experiments, we break ties encountered on \cref{line:beam-prune}
by comparing programs according to following sort key: 
$\tuple{d_{1}, \ldots, d_{N}}$ where $d_{i}$ is $i^{\text{th}}$ largest in the program with $N$ rules.
This comparison has the benefit that it will minimize lower degree terms as well, which may better guide search.
}
However, the pruning also robs breadth-first search of any guarantees. 
Increasing the beam size $B$ generally returns a lower-cost terminal state, but occasionally it may result in incorrectly pruning a good state and thus returning a higher-cost terminal state (as we see in our experiments).
We do recover exhaustive breadth-first search, which is guaranteed correct, when the beam size is large enough that \emph{no} pruning is done.

\newcommand{\beamsearch}[0]{\mysf{beam\_search}}
\newcommand{\thebeam}[0]{\mysf{beam}}

\begin{figure}
\begin{algorithmic}[1]\small
\Func{$\beamsearch(s_0, B)$}
    \State $\thebeam \gets [s_0]$
    \While{$\thebeam$}
        \State $\thebeam' \gets [\,]$
        \For{$s \in \thebeam$}
            \For{$a \in \mathcal{A}(s)$}
                \State $s' \gets \transition(s, a)$
                \State $\thebeam'$.append$(s')$
            \EndFor
        \EndFor
        \State $\thebeam \gets$ $B$ lowest-$\cost$ elements of $\thebeam'$ \label{line:beam-prune}
    \EndWhile
    \State \Return lowest-$\cost$ state ever to appear in $\thebeam$
\EndFunc
\end{algorithmic}
\caption{Beam search algorithm}
\label{alg:beam-search}
\end{figure}

\subsection{Monte Carlo Tree Search}
Monte Carlo Tree Search (MCTS) is a learning based algorithm \citep{kocsis2006bandit,coulom06mcts}, 
is considered a ``major breakthrough'' in computer Go \citep{swiechowski2021-mcts-survey,gelly2012grand-challenge,alpha-go}.
MCTS is an uninformed search algorithm,
which means that it is classified alongside well-known algorithms
such as breadth-first search, beam search, and iterative deepening search.
However, since MCTS is based on learning, it has some of the benefit of an informed search algorithm, such as A$^*$ \cite{hart68astar}, without the burden of designing or evaluating a heuristic function.  Essentially, it \emph{learns} its heuristic by sampling sequences of actions.
For our specific search problem, designing an A$^*$ heuristic that works well with all of our search actions (especially unfold and rule elimination) is challenging.
We discuss our choice further in \cref{sec:mcts-discussion}.

The application of MCTS to graph search is summarized in \cref{fig:search-algs}. For thorough surveys on MCTS, we refer the reader to the surveys by \citet{swiechowski2021-mcts-survey} and \citet{browne2012mcts-survey}.

\begin{figure}
\begin{algorithmic}[1]\small
\Func{$\mcts(s_0, C, R)$}
    \State mode $\gets$ explore
    \State \textbf{repeat $R$ times}: $\mcts'(s_0)$ \label{alg:mcts:repeat-loop}
    \State mode $\gets$ deploy  \label{alg:mcts:set-deploy}
    \State \Return $\mcts'(s_0)$
\EndFunc
\Func{$\mcts'(s)$}
    \LineComment{Terminal state, return cost and final state}
    \If{$s \in \mathcal{T}$} \Return $\langle s, \cost(s) \rangle$  \EndIf
    \State $a \gets \policy(s)$
    \LineComment{Transition to new state}
    \State $\langle s^*, c \rangle \gets \mcts'(\mysf{transition}(s, a))$
    \State $\Update(s, a, c)$
    \State \Return $\langle s^*, c \rangle$
\EndFunc
\Func{$\policy(s, \mathcal{A})$}
  \If{$\widehat{n}(s) = 0$} \Comment{Novel state, follow initial policy}
    \State \Return $\InitialPolicy(s)$
  \ElsIf{mode = explore}
    \State \Return $\argmin_{a \in \mathcal{A}}\ \frac{\widehat{c}(s, a)}{\widehat{n}(s,a)} - C \sqrt{\frac{ \log \widehat{n}(s) }{ \widehat{n}(s,a)} }$
  \Else  \Comment{Deploy mode}
    \State \Return $\argmax_{a \in \mathcal{A}}\ \widehat{n}(s, a)$  \label{alg:mcts:deploy-policy}
  \EndIf
\EndFunc
\Func{$\Update(s, a, c)$}
  \LinesComment{Update MCTS statistics after observing cost}
  \State $\widehat{c}(s,a) \ \texttt{+=}\ c$ ;\ $\widehat{n}(s,a) \ \texttt{+=}\ 1$ ;\ $\widehat{n}(s) \ \texttt{+=}\ 1$
\EndFunc
\end{algorithmic}
\caption{Search algorithm}
\label{fig:search-algs}
\end{figure}

MCTS searches by estimating the expected cost-to-go for taking action $a$ in a given state $s$, $\frac{\widehat{c}(s,a)}{\widehat{n}(s,a)}$ where $\widehat{c}(s,a)$ is the total cost of previous attempts of action $a$, and $\widehat{n}(s,a)$ is the total number of such attempts.  In order to \emph{learn} a \defn{policy} $\policy$ that maps states to actions, MCTS selects the action $a$ in state $s$ that minimizes the \defn{lower-confidence bound},
\begin{equation}
 \frac{\widehat{c}(s, a)}{\widehat{n}(s,a)} - C \sqrt{\frac{ \log \widehat{n}(s) }{ \widehat{n}(s,a)} } \label{eq:lcb}
\end{equation}
In state $s$, MCTS chooses the action $a$ that minimizes \cref{eq:lcb}.  This bound treats actions optimistically in the face of uncertainty: if an action in state $s$ has been under-explored, its cost \emph{might} be rather lower than the noisy average cost observed for it so far, and so MCTS may be willing to try it again.

The constant $C > 0$ is a tunable constant that controls the exploration--exploitation tradeoff.\footnote{In our experiments, we set $C$ to equal the degree of the initial program.  This is close to the theoretical requirement of an upper bound on the range (max - min) of the cost function.}  If $\widehat{n}(s,a) = 0$, the lower confidence bound is defined to be $-\infty$; thus, novel actions are always explored if there are any.
In this paper, we use MCTS as a batch search algorithm. That is why the top-level \text{MCTS} routine includes a repeat-loop (\cref{alg:mcts:repeat-loop}) and switches the policy into deployment mode (\cref{alg:mcts:set-deploy}).  Notice that when the policy is in deployment mode, it \emph{exploits} by selecting the most frequently explored action (\cref{alg:mcts:deploy-policy}).\footnote{It is also reasonable to use the action with the lowest estimated cost.  However, this choice is less stable in practice.}
Lastly, we note if MCTS is run for sufficiently long and the constant $C$ is set appropriately, it will converge to an optimal transformation sequence \citep{kocsis2006bandit}.

\paragraph{Initial Policy Design}
MCTS can be greatly sped up using the following strategy: on the first visit to a state (i.e., $\widehat{n}(s) = 0$), we redirect control to an \defn{initial policy} $\InitialPolicy$ rather than \emph{uninformed} exploration (that which results from following the lower confidence bound).  This results in a sensible initial value for the cost-to-go estimate.  For our initial policy, we randomly fold all rules until there are no more fold actions available.

\subsection{Refinements to the Search Graph}
\label{sec:search-refined}

In the sections describing each of the transforms, we discussed conditions for the transforms to be valid.  
The basic version of the search graph would simply say that all valid transforms from \cref{sec:transforms} are available at all times.  However, that would ignore some useful problem structure.  In this section, we propose two refinements to the search space: rule to-do lists and macro folding.  We validate their empirical utility in \cref{sec:ablations}.

\paragraph{Rule to-do list.}
Each of the transforms we consider is centered around a specific rule in the current version of the transformed program.  Applying transforms to rules $r$ and $r'$ in either order will get the same result if neither transform makes the other one impossible.  Thus, we consider transforms in a canonical order.  Each state will now consist of a program together with \defn{to-do list} of rules that can still be transformed.  The possible actions at that state consist of either removing the top rule $r$ from the list (declining to transform $r$) or applying a program transform (fold, unfold, or elimination) that is centered on $r$.  Applying such a transform may delete and/or add program rules, which are correspondingly deleted from the list and/or added at the bottom of the list.  (If the list is empty, no more actions are possible.)
This design reduces the branching factor by a factor of the number of rules, and improves the sharing of statistics at nodes in the MCTS search tree.  
Of course, a potential downside is that it makes the search tree deeper by a factor of the number of rules.\looseness=-1

\paragraph{Macro folding.}
Our most important refinement is to use macro folding actions.
These actions will take a given rule $r$ and completely fold it \emph{independently} from the main program allowing us to memoize it.
More precisely, macro-folding runs the program containing the single rule $r$ through search $\Prog'_r \gets \mysf{search}_{\text{fold-only}}([r])$, and then merges the solution into the main program, $(\Prog - r) \cup \Prog'_r$.
Macro folding provides an exponential reduction in the size of the search space because it allows any given rule to be optimized by folding independently of the other rules in the program.  Thus, if a given rule appears in multiple program variants,
we can re-use knowledge acquired from folding it in other contexts to fold it in the current context---analogous to memoizing the best folding sequence for each rule.
The macro folding action is implemented in our graph search instance as yet another action.\footnote{When we enable macro folding in our experiments, we disable the basic fold action since they are redundant.}
However, unlike the other transforms, macro folding is useful to memoize as it is reusable across many of the programs explored during search.\looseness=-1

\subsection{Discussion}
\label{sec:mcts-discussion}

Our goal in this work was to exhibit a working method (not necessarily the best one).  But since our search problem is just graph search, why not simply use a classical method like A$^*$ \citep{hart68astar}?  The challenge is in designing an admissible and effective A$^*$ heuristic.  The role of the heuristic is to approximate lookahead.  MCTS does not need a hand-designed heuristic because it instead performs lookahead by actual rollouts.  The average cost of these rollouts is still only an approximation of the optimal cost-to-go, because the rollouts use the current exploration policy---but it approaches the optimal cost-to-go as the algorithm continues to run.  MCTS has been previously used for graph search \cite{mcts-arch-search,Negrinho-2019}.\looseness=-1

Could an A$^*$ heuristic be designed in our setting?  There are many good search heuristics (e.g., \citet{gogate04treewidth}) in the special case where only folding actions are allowed.
However, for unfold and rule elimination, the heuristics are difficult to derive.  The challenge with these actions is that they are always uphill moves with delayed benefits: it is often the case that we require several unfolds, each increasing the \degree, followed by several (potentially tricky) folds.

\begin{table}[t]
\centering
\begin{adjustbox}{width=.85\columnwidth}
\begin{tabular}{lrrrr}
      & \multicolumn{2}{c}{avg rel degree}
      & \multicolumn{2}{c}{\% optimal} \\      

benchmark &  beam &  mcts &  beam &  mcts \\
\midrule
\hyperref[sec:bar-hillel]{bar-hillel}  &  1.00 &                           1.00 &                       100 &                    100 \\
\hyperref[sec:bilexical-labeled]{bilexical-labeled}   &                              0.97 &                           1.00 &                       90 &                    100 \\
\hyperref[sec:bilexical-unlabeled]{bilexical-unlabeled} &                              1.00 &                           0.99 &                       100 &                    90 \\
\hyperref[sec:chain-10]{chain-10}            &                              1.00 &                           1.00 &                       100 &                    100 \\
\hyperref[sec:chain-20]{chain-20}            &                              1.00 &                           1.00 &                       100 &                    100 \\
\hyperref[sec:chain-expect]{chain-expect}          &                              1.00 &                           1.00 &                       100 &                    100 \\
\hyperref[sec:cky+grammar]{cky+grammar}            &                              0.74 &                           0.68 &                       40 &                    40 \\
\hyperref[sec:cky3]{cky3}                &                              0.99 &                           0.99 &                       90 &                    90 \\
\hyperref[sec:cky4]{cky4}                &                              0.97 &                           0.96 &                       90 &                    80 \\
\hyperref[sec:edit]{edit}                &                              1.00 &                           0.99 &                       100 &                    90 \\
\hyperref[sec:hmm]{hmm}                 &                              1.00 &                           1.00 &                       100 &                    100 \\
\hyperref[sec:itg]{itg}                 &                              0.98 &                           0.95 &                       90 &                    60 \\
\hyperref[sec:path]{path}         &                              1.00 &                           1.00 &                       100 &                    100 \\
\hyperref[sec:semi-markov]{semi-markov}         &                              1.00 &                           1.00 &                       100 &                    100 \\
\hyperref[sec:split-head]{split-head}          &                              0.99 &                           0.99 &                       90 &                    90 \\
\end{tabular}
\end{adjustbox}
\caption{Experimental results for stress test experiments.  Each row is a class of 10 randomly constructed, semantically equivalent input programs.
 \begin{itemize*}
 \item \mysf{\%~optimal}: the percentage of the 10 random programs for which we find the optimal degree within the search budget ($R\!=\!300K$ iterations for MCTS, $B=1000$ for beam search). In many rows, this is 100\%.  However, in some recursive cases with two or more recursive subgoals, the randomly applied unfolds make the programs very big, which makes them difficult to optimize.  Both methods performed poorly on the \hyperref[sec:cky+grammar]{cky+grammar} benchmark, which is by far the longest program we consider as it contains 35 rules in its original form.  
 \item Average relative degree: The relative degree achieved by search, averaged over the 10 random programs.  We see that this metric follows the same trend as \% optimal.
 \item Overall, we see a small but consistent improvement of $\beamsearch$ over $\mcts$ (under both metrics), except on bilexical-labeled.
 \end{itemize*}}\label{tab:results}
 \vspace{-\baselineskip}
\end{table}

\newcommand{\n}[0]{--}
\newcommand{\y}[0]{+}
\begin{table}[t]
\centering
\begin{adjustbox}{width=.8\columnwidth}
\begin{tabular}{llccrr}
search & $B$ & todo & macro &  avg rel deg &  \% optimal \\
\midrule
beam & 100  & \n & \n &             96 &     87 \\
     &      & \y & \n &             93 &     75 \\
     &      & \y & \y &             97 &     93 \\
     \cline{2-6}
     & 1000 & \n & \n &             90 &     74 \\
     &      & \y & \n &             94 &     77 \\
     &      & \y & \y &             98 &     93 \\ 
\hline
mcts &      & \n & \n &             97 &     83 \\
     &      & \y & \n &             96 &     80 \\
     &      & \y & \y &             97 &     89 \\
\end{tabular}
\end{adjustbox}
\caption{Ablation of search-space refinements.  Like \cref{tab:results}, we show \% optimal and average relative degree, except here we average together all benchmarks to see an overall picture.  
The rows are labeled by their search method and whether that search method operates on a search graph with a to-do list and/or macro folding.  Overall, we see that the proposed refinements improve overall performance under both metrics, except that smaller beams are disadvantaged by the increased depth of the to-do list refinement.
When the macro folds are added to the small beam, performance hits the same peak as the system with a larger beam.
\vspace{-\baselineskip}
}
\label{tab:ablation}
\end{table}

\section{Experiments}
\label{sec:experiments}

The goal of this paper was to devise a system for automatically improving typical dynamic programming problems.  To evaluate whether we achieved this goal, we devised a set of \emph{unit tests} and \emph{stress tests} to see how well our proposed approach works.

\paragraph{Unit tests.}
Our unit tests include most of the faux pas mentioned in \cref{sec:intro}; the precise set of programs is provided in \cref{app:programs}.
However, recovering these instances is relatively easy as they typically only require a few fold transformations.  Thus, they are not a good stress test for our automated system. In all of our test cases, we know the optimal $\degree$, and we have verified that both MCTS (with $R=100$ iterations) and beam search (with a beam of size $B=10$) successfully find it within a few seconds.\looseness=-1

\paragraph{Stress tests.}
The inspiration for our stress tests is to imagine that a \naive programmer produces a suboptimal program.  For example, they may have chosen the poor variable-elimination order in \cref{ex:z:bad-fold}, and, now, we would like to ``undo'' their handiwork via unfold, elimination, and fold.  We operationalize such a \naive programmer by imaging that they have applied a sequence of random folds and unfolds to a starting point program.
More precisely, for each program $\Prog_0$ in our unit test suite with known optimal $\degree$ $d^*$, we generate an inefficient variant $\Prog_1$ for search to improve.  The variant $\Prog_1$ is generated by applying a random sequence of transformations to $\Prog_0$, but we reject $\Prog_1$ if the optimal degree $d^*$ can be ``trivially'' achieved by applying the greedy fold-only algorithm to $\Prog_1$.
The results of this experiment are summarized in \cref{tab:results}.\looseness=-1

We compare the search algorithms according to the percentage of stress tests that they are able to solve optimally.  We also consider the \defn{relative degree}, i.e., the fraction of the possible improvement that was actually achieved:
$\frac{\degree(\Prog_1) - \degree(\Prog)}{\degree(\Prog_1) - d^*}$.

\paragraph{Ablation analysis.}\label{sec:ablations}
We explore the utility of our propose search space refinements (\cref{sec:search-refined}) in \cref{tab:ablation}.

\section{Conclusion}
We have presented a system for automatically analyzing and improving dynamic programming algorithms. Expressing those algorithms in the Dyna programming language allows us to successively apply program transformations introduced by \citet{eisner-blatz-2007}.
We showed that Monte Carlo tree search and beam search allows us to automatically discover asymptotically faster algorithms.\looseness=-1

\section*{Acknowledgments}
We wish to thank Matthew Francis-Landau, Ran Zmigrod, and the anonymous reviewers for their helpful suggestions that improved this paper.

\bibliographystyle{acl_natbib}
\bibliography{main}

\appendix
\onecolumn
\section{Programs Used for Experiments}
\label{app:programs}

\subsection{Chain Structures (chain-10, chain-20)}
\label{sec:chain-10}\label{sec:chain-20}
Chain structures are common in NLP as they model interaction between adjacent words or label in a sequence (e.g., \citet{lafferty01crf}).
\\
\noindent The chain-N programs are adaptations of \cref{ex:z:dyna},
\begin{dynaex}[firstnumber=1]{}
z += w(Y1,Y2) * w(Y2,Y3) * w(Y3,Y4) * w(Y4,Y5).
\end{dynaex}
which is for the specific case of length 4, to have length N.
The original program's degree is N.
The optimal degree for chains is 2 regardless of N.

\subsection{Projective Dependency Parsing}

This program performs a version of CKY (\cref{sec:cky}) where the non-terminals have been lexically annotated with their head word.
In contrast to CKY's $\bigo{n^3}$ runtime, the original presentation of this algorithm ran in $\bigo{n^5}$ \citep{collins-1996-new}. 
In subsequent work \citet{eisner-satta-1999-efficient} gave a faster version of the algorithm that runs in $\bigo{n^4}$.\looseness=-1

\subsubsection{Bilexical Unlabeled (bilexical-unlabeled)}
\label{sec:bilexical-unlabeled}

\begin{dynaex}[firstnumber=1]{}
phrase(I,H,K) += phrase(I,H,J) * phrase(J,H',K) * score(H,H',left).
phrase(I,H',K) += phrase(I,H,J) * phrase(J,H',K) * score(H,H',right).
phrase(I,I,K) += word(I,I,K).
goal += phrase(0,_,N) * len(N).
input word(_,_,_); len(_); score(_,_,_).
output goal
\end{dynaex}

\noindent Degree: 5. Optimal: 4.

\subsubsection{Bilexical Labeled (bilexical-labeled)}
\label{sec:bilexical-labeled}
Extends bilexical-unlabeled with labels (i.e., grammar relations).

\begin{dynaex}[firstnumber=1]{}
% 0 right children so far
rconstit(X,H,I,K) += rewrite(X,H) * word(H,I,K).
% add right child
rconstit(X,H,I,K) += rewrite(X,H,Y,H,Z,H') * rconstit(Y,H,I,J) * constit(Z,H',J,K).
% 0 left children so far
constit(X,H,I,K) += rconstit(X,H,I,K).
% add left child
constit(X,H,I,K) += rewrite(X,H,Y,H',Z,H) * constit(Y,H',I,J) * constit(Z,H,J,K).
goal += constit(s,H,0,N) * len(N).
input word(_,_,_); len(_); rewrite(_,_,_,_,_,_).
output goal.
\end{dynaex}

\noindent Degree: 8. Optimal: 7.

\subsubsection{Split-Head-Factored (split-head)}
\label{sec:split-head}
References: \cite{johnson-2007-transforming,eisner-blatz-2007}

\begin{dynaex}[firstnumber=1]{}
goal += x(0,_,N) * len(N).
% words are "duplicated" as in Johnson (2007).
l(I,K) += word(left,I,K).
r(I,K) += word(right,I,K).
% (I, K) is a span with K as the head
l(I,K) += x(I,V,J) * l(J,K) * score(left, V, K).   % V -> K
% (I, K) is a span with I as the head
r(I,K) += r(I,J) * x(J,V,K) * score(right, V, I).  % I <- V
% J is the head of l(I, J) and J is the head of r(J, K)
x(I,J,K) += l(I,J) * r(J,K).
input word(_,_,_); score(_,_,_). len(_).
output goal(_).
\end{dynaex}

\noindent Degree: 4, Optimal: 3.

\subsection{CKY}
\label{sec:cky}
The following programs are variants of CKY \cite{cocke-schwartz1970,younger67parsing,kasami65cky,lange2009cnf,eisner-blatz-2007,10.5555/1623611.1623625,tomita2012generalized,johnson-1989-computational,baker79trainable,jelinek1985markov}.  We briefly discussed CKY in \cref{ex:cky} of the main text.

\subsubsection{CKY (cky3)}
\label{sec:cky3}
\begin{dynaex}[firstnumber=1]{}
beta(X,I,K) += gamma(X,Y,Z) * beta(Y,I,J) * phrase(Z,J,K).
beta(X,I,K) += gamma(X,Y) * beta(Y,I,K).
beta(X,I,K) += gamma(X,Y) * word(Y,I,K).
z += beta(root, 0, N) * len(N).
input word(W,I,K); len(N); gamma(X,Y,Z); gamma(X,Y)
output z
\end{dynaex}

\noindent Degree: 6, Optimal: 5.

\subsubsection{CKY with 4-ary Productions (cky4)}
\label{sec:cky4}

The following program implements \citet{10.5555/1623611.1623625}'s algorithm where the grammar can have a production rule of length up to 4.  It is a simple modification to CKY4.  We just add the following rule, and input declaration.
\begin{dynaex}{}
phrase(X,I1,I4) += gamma(X,Y1,Y2,Y3) * phrase(Y1,I1,I2) * phrase(Y2,I2,I3) * phrase(Y3,I3,I4).
input gamma(X,Y1,Y2,Y3)
\end{dynaex}

\noindent Degree: 7, Optimal: 6.

\subsubsection{CKY with a Fixed Grammar (CKY+grammar)}
\label{sec:cky+grammar}
For CKY+grammar, we use CKY3, but we remove the input declaration for $\gamma$ and declare the following grammar for the program to specialize to.
\begin{dynaex}{}
gamma("S", "NP", "VP") += 1.0.
gamma("NP", "Det", "N") += 1.0.
gamma("NP", "NP", "PP") += 1.0.
gamma("VP", "V", "NP") += 1.0.
gamma("VP", "V") += 1.0.
gamma("VP", "VP", "PP") += 1.0.
gamma("PP", "P", "NP") += 1.0.
gamma(@{\color{black}"<.>"}@, ".") += 1.0.
gamma("NP", "Papa") += 1.0.
gamma("N", "caviar") += 1.0.
gamma("N", "spoon") += 1.0.
gamma("N", "fork") += 1.0.
gamma("N", "telescope") += 1.0.
gamma("N", "boy") += 1.0.
gamma("N", "girl") += 1.0.
gamma("N", "baby") += 1.0.
gamma("N", "man") += 1.0.
gamma("N", "woman") += 1.0.
gamma("N", "moon") += 1.0.
gamma("N", "cat") += 1.0.
gamma("V", "ate") += 1.0.
gamma("V", "saw") += 1.0.
gamma("V", "fed") += 1.0.
gamma("V", "said") += 1.0.
gamma("V", "jumped") += 1.0.
gamma("P", "with") += 1.0.
gamma("P", "over") += 1.0.
gamma("P", "under") += 1.0.
gamma("P", "above") += 1.0.
gamma("P", "below") += 1.0.
gamma("P", "on") += 1.0.
gamma("P", "in") += 1.0.
\end{dynaex}

\noindent Degree: 6, Optimal: 3.

\subsubsection{Inversion Transduction Grammars (itg)}
\label{sec:itg}

Inversion transduction grammars were introduced by \citet{wu-1996-polynomial}, who gave
an $\bigo{n^7}$ algorithm, which was later sped up to $\bigo{n^6}$ by \newcite{huang-etal-2005-machine}.  The model is one that simultaneously parses a pair of related sentences---typically a target language sentence and a source language sentence as in machine translation.  The model allows for a restricted form of syntactic reordering of phrases called inversion.

\begin{dynaex}[firstnumber=1]{}
constit(A,I,K,I',K') += word(X,I,K) * word'(X',I',K') * transduce(A,X,X').
constit(A,I,K,I',K') += constit(B,I,J,I',J') * constit(C,J,K,J',K') * rewrites(A,B,C).
constit(A,I,K,I',K') += constit(B,J,K,I',J') * constit(C,I,J,J',K') * rewrites_inv(A,B,C).
goal += constit("A",0,M,0,N) * lenM,N).
input word(_,_,_); word'(_,_,_); transduce(_,_,_); rewrites(_,_,_); rewrites_inv(_,_,_); len_,_).
output goal.
\end{dynaex}

\noindent Degree: 9, Optimal: 8.

\subsection{Edit Distance (edit)}
\label{sec:edit}
The following program implements a weighted generalized monotonic alignment between two sequences \dd{word} and \dd{word'}.
It is essentially the well-known Levenstein distance \citep{Levenshtein}.

\begin{dynaex}[firstnumber=1]{}
align(0,0) min= 1.
align(J,J') min= align(I,I') * word(W,I,J) * word'(W',I',J') * score(W,W').
align(I,J') min= align(I,I')               * word'(W',I',J') * score(@$\varepsilon$@,W').
align(J,I') min= align(I,I') * word(W,I,J)                   * score(W,@$\varepsilon$@).
goal min= align(N,N') * len(N) * len'(N').
input word(_,_,_); word'(_,_,_); score(_,_); len(_); len'(_).
output goal.
\end{dynaex}

\noindent Degree 6, Optimal: 4.

\subsection{Bar-Hillel Construction (bar-hillel)}
\label{sec:bar-hillel}
The following program implements a parser for (weighted) intersection of a context-free parser and a bigram model on the part-of-speed sequences.  It is essentially \citet{bar-hillel61grammars}'s construction of a context-free language that accepts the intersection of a regular language and a context-free language.

\begin{dynaex}[firstnumber=1]{}
goal += beta(0,_,root,_,N) * len(N).
beta(I,A,X,D,K) += beta(I,A,Y,B,J) * beta(J,C,Z,D,K) * gamma(X,Y,Z) * bigram(B,C).
beta(I,X,X,X,K) += tag(X,W) * word(W,I,K).
input len(_); word(_,_,_); bigram(_,_); gamma(_,_,_); tag(_,_).
output goal.
\end{dynaex}
\noindent Degree 10, Optimal: 8.

\subsection{Expectations under a Linear-Chain Conditional Random Field (chain-expect)}
\label{sec:chain-expect}
This example implements the inside-outside speedup \citep{li-eisner-2009} for computing the expectation of an additively decomposable function $f: \vS \times \vS' \to \mathbb{R}^d$ over randomly drawn sequences from a weighted graph (e.g., a conditional random field \citep{lafferty01crf}).  The graph is specified as a collection of weights \dd{w}, as well as \dd{start} and \dd{end} nodes.
The relations $\alpha$ and $\beta$ implement the forward-backward algorithm (discussed in \citet{rabiner1989tutorial}), and \dd{z} is the normalization constant of the distribution.  The expectation of the $i^{\text{th}}$ dimension of \dd{f} is \dd{fbar(\vI)/z}.

\begin{dynaex}[firstnumber=1]{}
% forward algorithm
alpha(S) += start(S).
alpha(S') += alpha(S) * w(S,S').
% backward algorithm
beta(S) += end(S).
beta(S) += w(S,S') * beta(S').
% normalization constant
z += alpha(S) * end(S).
% unnormalized expected value via inside-outside speedup
fbar(R) += alpha(S) * w(S,S') * beta(S') * r(S,S',R).
input w(_,_); r(_,_,_); start(_); end(_).
output fbar(_). z.
\end{dynaex}
\noindent Degree 3, Optimal: 3.

\subsection{Hidden Markov Models (hmm)}
\label{sec:hmm}
Hidden Markov models (HMMs) are the generative and locally normalized analogue of CRFs, which are discussed in \cref{sec:chain-expect}.
\citet{rabiner1989tutorial} provides a classic tutorial.

\begin{dynaex}[firstnumber=1]{}
v(0,start) += 1.
v(T',Y') += v(T,Y) * emission(Y,X) * transition(Y,Y') * obs(T,X,T').
goal += v(N,stop) * len(N).
input obs(_,_,_); len(_); emission(_,_); transition(_,_).
output goal.
\end{dynaex}
\noindent Degree 5, Optimal: 4.

\subsection{Semi-Markov Model (semi-markov)}
\label{sec:semi-markov}
The a semi-Markov model \citep{sarawagi04semi} generalizes a Markov model to score spans rather than individual words. 
In terms of runtime, one can perform inference in a Markov model in $\bigo{n}$ time (omitting dependence on the number of tags). 
In contrast, inference in a semi-Markov model takes $\bigo{n^2}$.

\begin{dynaex}[firstnumber=1]{}
beta(start, 0) += 1.
beta(Y, J) += beta(X, I) * transition(X, Y) * chunk(Y, I, J).
goal += beta(_, N) * len(N).
input transition(_,_); chunk(_, _, _); len(_).
output goal.
\end{dynaex}
\noindent Degree 4, Optimal: 3.

\subsection{Most Probable Path (path)}
\label{sec:path}
\cref{ex:viterbi} of the main text briefly discussed the most-probable path algorithm \citep{viterbi67error}.  We give a slightly more general version here that finds the most probable path from a set of \dd{start} states to a set of \dd{end} states.

\begin{dynaex}[firstnumber=1]{}
v(S) max= start(S).
v(S') max= v(S) * w(S, S').
goal max= v(S) * stop(S).
input w(_,_); start(_); stop(_).
output goal.
\end{dynaex}
\noindent Degree 2, Optimal: 2.

\newpage
\section{Pseudocode for Program Transformations}
\label{app:transform-pseudocode}

In this section, we will make more extensive use of manipulations of terms.
Terms can be equated with---or, matched against---other terms via \defn{structural equality constraints} \citep{herbrand1930thesis,robinson65unification,martelli1982efficient,knight89unification}, which are denoted by the equality
operator, e.g., \dd{f(\vX) = f(3)}.  Systems of structural equality constraints have a unique minimal solution (up to variable renaming) when a solution exists, known as the most general unifier.  For example, \dd{f(\vY,\vZ) = f(g(\vX),4)} has the solution \dd{\{\vY $\mapsto$ g(\vX), \vZ $\mapsto$ 4\}}, whereas \dd{f(\vX,g(\vX)) = f(3,g(4))} has no solutions (is unsatisfiable).  We assume access to a subroutine $\Unify$ that can find a \defn{substitution} mapping $\vtheta$ to that makes the terms equal, or returns $\vtheta = \emptyset$ if no substitution exists. For example, $\unify{\dd{f(\vY,\vZ)}}{\dd{f(g(\vX),4)}} \mapsto \vtheta \!=\! \dd{\{\vY \mapsto g(\vX), \vZ \mapsto 4\}}$.
We can apply the substitution with $\subst{ \dd{f(\vY,\vZ)} }{ \vtheta } \!=\! \dd{f(g(\vX),4)}$.
We will make use of the following utility method: $\fresh{x} \mapsto x'$ which returns a term $x'$, which denotes the same set as the term $x$, but has distinct variable names, $\vars{x} \!\cap\! \vars{x'} \!=\! \emptyset$.  This operation is useful to prevent variable naming conflicts.  For example, $\fresh{\dd{f(g(\vX),\vX)}} \!=\! \dd{f(g(\vX_*),\vX_*)}$ where $\dd{\vX_*}$ is a novel variable name.

\vspace{20pt}

\begin{algorithmic}[1]
\Func{$\fold(\Prog, i, \valpha)$}  \label{alg:fold}
\LinesComment{input: rule index $i$, subgoal indices to fold into a new subgoal $\valpha$.}
\State $(\dhead \ \dopluseq\ \db_1 \dotimes \ldots \dotimes \db_K) \gets \Prog_i$
\State $\vbeta \gets \{1, \ldots, K\} \smallsetminus \valpha$ \Comment{Remaining factors}
\State $\{ \vX_1, \ldots, \vX_K \} \gets \vars{\db_{\valpha}}\,\smallsetminus (\vars{\db_{\vbeta}} \cup \vars{\dhead})$
  \LinesComment{Generate a new relation with a unique name, provided by the $\gensym()$ utility method.  Note that the ordering of the arguments is arbitrary, but it is important for it to be used consistently.}
  \State $\texttt{gen}_{\star} \gets \gensym()$
  \State $h' \gets \dd{gen_{\star}}(\vX_1, \ldots \vX_K)$
  \State $\Prog' \gets \Prog$   \Comment{Copy rules}
  \State $\Prog'_i \gets \left(\dhead\ {\dopluseq}\ h' \dotimes \dprod_{j \in \vbeta}\,\db_j \right)$ \label{alg:unfold:front}
  \State $\Prog'$.append$\left( h'\ \dopluseq\ \dprod_{j \in \valpha}\,\db_j \right)$   \Comment{Add new rule that defines $h'$}
\State \Return $\Prog'$
\EndFunc
\end{algorithmic}

\vspace{20pt}

\begin{algorithmic}[1]
\Func{$\unfold(\Prog, i, k)$} \label{alg:unfold}
\State $(\dhead \ \dopluseq\ \db_1 \dotimes \ldots \dotimes \db_K) \gets \Prog_i$
\State $\Prog' \gets$ remove$(\Prog, i)$
\For{$j = 1 \ldots |\Prog|$}
  \LinesComment{In the case of recursion, we rename variables in $s'$ to avoid variable-name collisions.}
  \State $s \gets \fresh{\Prog_j} \textbf{ if } i = j \textbf{ else }\Prog_j$
  \LinesComment{Solve for a substitution to make the head match \dhead}
  \State $\vtheta \gets \unify{\head{s}}{\db_k}$
  \If{$\vtheta = \emptyset$} \textbf{continue} \EndIf
  \LinesComment{Copy rule body}
  \State $r' \gets (\dhead \ \dopluseq\
    \db_1 \dotimes \ldots \dotimes \db_{k-1} \dotimes\,
    \body{s} \,\dotimes\, \db_{k+1} \dotimes \ldots \dotimes \db_{K})$
  \LinesComment{Apply substitution; copy rule}
  \State $\Prog' \gets \Prog' \cup \{ \fresh{\subst{r'}{\vtheta}} \}$
\EndFor
\State \Return $\Prog'$
\EndFunc
\end{algorithmic}

\newpage
\begin{algorithmic}[1]
\Func{$\eliminate(\Prog, s)$}
\LinesComment{Transform assumes that all rules are range-restricted.}
\State $\Prog \gets \linearize(\Prog, \head{s})$  \label{line:linearize-call}
\State $\Prog' \gets \texttt{[]}$
\For{$r' \in \Prog$}
    \State $r \gets \fresh{r'} \textbf{ if } r' \textbf{ is } s \textbf{ else } r'$
    \State count = 0
    \State $(\dhead \ \dopluseq\ \db_1 \dotimes \ldots \dotimes \db_k) \gets r$
    \For{$i = 1 \ldots k$}
        \State $\vtheta \gets \unify{\head{s}}{\db_i}$
        \If{$\vtheta \ne \emptyset$}
            \State count += 1
            \State $r' \gets (\dhead \,\dopluseq\, \db_1 \dotimes \ldots \dotimes \db_{i-1} \dotimes\ \body{s}\ \dotimes\, \db_{i+1} \dotimes \ldots \dotimes \db_k)$
            \State $\Prog'$.append$(\fresh{\subst{r'}{\vtheta}})$
        \EndIf
    \EndFor
    \State \textbf{assert } count $> 1$ \Comment{ensured by linearize on \cref{line:linearize-call}}
    \If{$r' \textbf{ is } s$} \textbf{continue} \EndIf
    \State $\Prog'$.append(r)
\EndFor
\State \Return $\Prog'$
\EndFunc
\end{algorithmic}

\noindent The $\linearize$ utility method transforms a program with repeated subgoals.
For example, it transforms
\begin{dynaex}[firstnumber=1]{}
goal += f(X) * g(X,Y) * f(Y).
\end{dynaex}
into the following equivalent program that that does not repeat the \dd{f} subgoal---or, more precise, it does not have any pair of subgoals that unify.
\begin{dynaex}{}
goal += f(X) * g(X,Y) * gen(Y).
gen(Y) += f(Y).
\end{dynaex}
This transformation is useful as pre-processing in the $\eliminate$ function, which assumes the there are no repeated subgoals.  If we do not want the entire program to be linearized, but just sufficiently linearize to $\eliminate$ a specific rule, we can specify what term we want to be linear with respect to.  This is used in \cref{line:linearize-call} of the $\eliminate$ pseudocode.
\newline

\newcommand{\dz}[0]{\dd{z}}

\begin{algorithmic}[1]
\Func{$\mysf{linearize}(\Prog, \dz)$}
\State $\Prog' \gets \Prog$ \Comment{copy rules}
\For{$i = 1 \ldots |\Prog|$}
    \State $(\dhead \,\dopluseq \, \db_1 \dotimes \ldots \dotimes\, \db_K) \gets \Prog_i)$
    \For{$j = 1 \ldots K$}
        \If{ $\unify{\db_j}{\dz} = \emptyset$} \textbf{continue} \EndIf
        \For{$k = j\!+\!1 \ldots K$}
            \If{$\unify{\db_j}{\db_k} = \emptyset$} \textbf{continue} \EndIf
            \LinesComment{$\fold$ subgoal $k$ out of rule $i$}
            \State $\db_k(\vX_1, \ldots, \vX_L) \gets \db_k$
            \State $\dd{gen_*} \gets \gensym()$
            \LinesComment{replace $\db_k$ with the $\dd{gen_*}$ subgoal so that it won't appear twice.}
            \State $\Prog'_{i} \gets \fresh{\dhead \ \dopluseq \ \db_1 \dotimes \ldots \dotimes\, \dd{gen_*}(\vX_1, \ldots, \vX_L)\, \dotimes \ldots \dotimes\, \db_K}$
            \State $\Prog'$.append$(\fresh{\dd{gen_*}(\vX_1, \ldots, \vX_L)\ \dopluseq\ \db_k(\vX_1, \ldots, \vX_L)})$

        \EndFor
    \EndFor
\EndFor

\State \Return $\Prog'$
\EndFunc
\end{algorithmic}

\end{document}